# MosquIoT: A System Based on IoT and Machine Learning for the Monitoring of *Aedes aegypti* (Diptera: Culicidae)

Javier Aira , Teresa Olivares , Francisco M. Delicado  , and Darío Vezzani

*Abstract*—Millions of people around the world are infected with mosquito-borne diseases each year. One of the most dangerous species is *Aedes aegypti*, the main vector of viruses such as dengue, yellow fever, chikungunya, and Zika, among others. Mosquito prevention and eradication campaigns are essential to avoid major public health consequences. In this respect, entomological surveillance is an important tool. At present, this traditional monitoring tool is executed manually and requires digital transformation to help authorities make better decisions, improve their planning efforts, speed up execution, and better manage available resources. Therefore, new technological tools based on proven techniques need to be designed and developed. However, such tools should also be cost-effective, autonomous, reliable, and easy to implement, and should be enabled by connectivity and multi-platform software applications. This paper presents the design, development, and testing of an innovative system named "MosquIoT". It is based on traditional ovitraps with embedded Internet of Things (IoT) and Tiny Machine Learning (TinyML) technologies, which enable the detection and quantification of *Ae. aegypti* eggs. This innovative and promising solution may help dynamically understand the behavior of *Ae. aegypti* populations in cities, shifting from the current reactive entomological monitoring model to a proactive and predictive digital one.

## I. Introduction

One of the most dangerous animals on earth is a tiny insect, the mosquito. More than 500 million people are infected with mosquito-borne diseases every year, with more then 3 million of them dying from these infections [1]. In addition, currently, half of the world's population, about 3.5 billion people, are at risk of infection from mosquito-borne diseases. Mosquitoes can be anthropophilic, carriers of virus, protozoans, and nematodes, making them highly effective and mobile agents for transmitting dangerous diseases among human populations [1]. One of the most dangerous species of anthropophilic mosquitoes is the *Aedes aegypti*, the main vector of viruses such as dengue, yellow fever, chikungunya, and Zika [2], [3], [4]. For example, before 1970, only 9 countries had experienced dengue epidemics, but the disease is now endemic in more than 100 countries in Africa, the Americas, the Eastern Mediterranean, South-East Asia and the Western Pacific. The Americas, South-East Asia and Western Pacific are the most seriously affected regions [2], [5]. *Ae. aegypti* may satisfy all of its biological requirements in human habitats, and may even develop in arid urban settings as a result of man-made humid conditions prevailing in cities. The essential resource for the proliferation of *Ae. aegypti* is water stored in artificial containers in urban areas, where they develop through their immature stages. Therefore, the most effective and efficient preventive measures are those aimed at eliminating water storage containers. *Ae. aegypti* has typically domestic habits and, although it has urbanized, the species has also been found in peri-urban, rural, and even wild habitats [6].

As part of current entomological surveillance campaigns, the reliable and proven ovitrap method is used [7] (Fig. 1, right) for detecting and quantifying *Ae. aegypti* eggs. This method is extremely sensitive and efficient in generating early alerts [8], [9], [10], [11], [12], [13]. Entomological surveillance campaigns are planned and executed by government agencies. To determine the existence of eggs, ovitraps are collected 7 days after they are set up. Samples should be gathered carefully and taken to the labs, ensuring they arrive there in appropriate conditions. These processes rely on good practices by the personnel involved, since, in addition to careful transportation, samples should also be moved under certain temperature and humidity conditions to ensure their preservation. Given these circumstances, the potential benefit and value of an automatic system that would provide real-time information about the behavior of *Ae. aegypti* populations in cities is evident, in that it would avoid the logistics involved in transporting samples to the labs, the lab analyses (preparation and reading times) and associated costs. A mass roll-out of such a solution would allow for the automatic building of dynamic risk maps and the development of predictive models to provide the authorities

This work was supported under PID2021-123627OB-C52 project, which funded by MCIN/AEI/10.13039/501100011033 and by European Regional Development Fund (ERDF), "A way to make Europe".

Javier Aira is a Ph.D. student at the International Doctoral School of the University of Castilla-La Mancha, 02071 Albacete, Spain (e-mail: jorgejavier.aira@alu.uclm.es).

Teresa Olivares is an Associate Professor with the Department of Computing Systems at the University of Castilla-La Mancha, 02071 Albacete, Spain (e-mail: teresa.olivares@uclm.es).

Francisco M. Delicado is an Associate Professor at the Department of Computer Engineering at the University of Castilla-La Mancha, 02071 Albacete, Spain (e-mail: francisco.delicado@uclm.es).

Darío Vezzani is a member of the National Scientific and Technical Research Council (CONICET), at the Multidisciplinary Institute on Ecosystems and Sustainable Development (UNCPBA–CICPBA), B7000 Tandil, Buenos Aires, Argentina (e-mail: dvezzani@gmail.com).

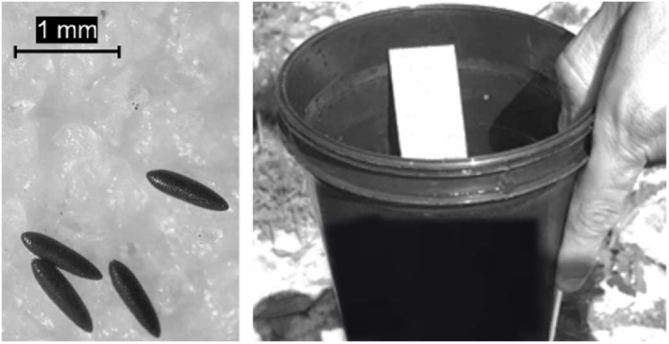

Fig. 1. Left: *Ae. aegypti* eggs. Right: Traditional ovitrap.

with digital tools that enhance their campaigns to prevent and contain potential epidemic outbreaks of these diseases.

Regarding the field of technological mosquito monitoring and surveillance systems, there is diverse literature on embedded trap systems, some of which are at the investigation stage, while others are commercially available. An analysis of the published research studies reveals proposals based on the use of sensors to classify insects by their flight sounds [14], acoustic traps for the surveillance of *Ae. aegypti* during rear-and-release operations [15], vision-based counting and recognition systems for flying insects [16], [17], and mosquito counting systems based on optical sensing [18], [19]. At a more advanced stage, we may also find products or solutions which are more consolidated at the industrial level. For example, Vectrack is a complex device able to identify mosquito species, sex, and age using optoelectronic sensors [20]. Additionally, Bzigo is a mosquito laser detector using a new flying insect localization system that pinpoints the mosquito´s location, even in the dark [21]. Finally, BG-Sentinel mimics convection currents created by the human body, employs attractive visual cues, and releases artificial skin emanations to capture selected mosquito species [22]. The above-mentioned systems are complex and expensive, limiting their scalability.

Considering the substantial impact of *Ae. aegypti* on health, the goal of this research study was to propose an improvement and upgrade to the current entomological surveillance systems, implementing a solution based on disruptive technologies such as the IoT and TinyML. This solution primarily comprises an IoT-based system called "MosquIoT", which was designed and developed under the functional concept of traditional ovitraps. The goal was to digitize this traditional and efficient method, whose name may be understood as a tool to "hunt" mosquitos but which, in fact, works as a sensor of the female mosquito's oviposition activity. Accordingly, the proposed system would enable the automatic detection and counting of eggs laid by *Ae. aegypti*, remotely sending such information to a cloud platform through several IoT communication links, for viewing, storage and subsequent examination. This study also focuses on determining the effectiveness and efficiency of the MosquIoT system. To do so, we present a first functional laboratory prototype, showcasing its technical and functional characteristics and the empirical evidence of its correct functioning in detecting *Ae. aegypti* eggs through actual samples used for the system training and validation.

The rest of the document is structured as follows: Section II delves into *Ae. aegypti* oviposition behavior and delimits the geographic area of application of this paper. Section III describes the system's global architecture, with special emphasis on the interplay between its various building blocks. Section IV provides a detailed description of MosquIoT. Section V contains a technical and functional description of a Mobile App associated with MosquIoT for its setup on the installation site. Section VI describes the development of the IoT platform, where the information on the evidence collected using MosquIoT will be concentrated, stored and displayed. Section VII describes the experimental validation processes that allowed us to demonstrate the technical and functional feasibility of MosquIoT. Finally, Section VIII contains the conclusions and future projects.

## II. AEDES AEGYPTI: OVIPOSITION BACKGROUND

The breeding habitats of *Ae. aegypti* are primarily man-made. These include old tires, flower vases, bottles, drinking troughs, or containers of any kind found in various urban areas, such as wasteland, graveyards, dumpsites, and dwellings. When the conditions are suitable, *Ae. aegypti* tend not to travel far from aquatic sites, where the mosquito develops its immature stages. However, under certain conditions, dispersion up to 800 meters has been observed [23], [24]. When a female *Ae. aegypti* completes its feeding, she lays individual, disperse eggs in different places, which is known as "skip oviposition". The female is attracted to dark or rigid wall-shaded recipients and prefers relatively clean waters with little organic content [25]. As for reproduction, this species lays eggs a few millimeters above fresh water, where larvae will pupate and subsequently emerge as adults [26]. In nature, many factors contribute to the choice of the egg-laying site: the amount of food available in the body of water, the number of larvae already present, the temperature, the humidity, the amount of light, and, primarily, the fresh nature of the water [6], [26], [27]. During oviposition, *Ae. aegypti* females may lay from 50 to 150 tiny eggs (size: $\approx$ 0.8 mm). Such eggs are laid in the recipient walls above the water level. When the recipient is again filled with water, the eggs sink and hatch. A group of eggs will hatch every time the water level rises in the recipient. In this way, hatching occurs as a staggered process to ensure survival even under unfavorable conditions (for instance, in times of drought). At the time of oviposition, eggs are white and almost transparent and become darker when in contact with air (Fig. 1, left).

In the Americas, the southern limit for dengue spread is the Province of Buenos Aires (Argentina). Each epidemic over the last decade has seen this limit shift southward. Saladillo is currently the southernmost location (in Argentina) with indigenous cases of dengue. Accordingly, this study is conducted in the city of Tandil, Province of Buenos Aires, Argentina. Tandil is a medium-sized city located some 200 km south of the current limit of dengue spread, where *Ae. aegypti* was not detected until early 2019. Upon such findings, several entomological studies led by Dr. Vezzani, provided empirical evidence that *Ae. aegypti* had settled in the city [12]. Against this backdrop, and for purposes of this study, a significant number of *Ae. aegypti* eggs were available from ovitraps set up in the city of Tandil. The geographic area of application of this study is Argentina. However, in the future, this scope may be expanded to other countries in America or other continents, as the sites, times, and seasons in which *Ae. aegypti* females lay eggs follow foreseeable behavioral patterns [25]. Such information is meaningful to execute prevention and surveillance actions through digital solutions such as that proposed in this paper.

## III. MOSQUIOT ARCHITECTURE

As mentioned, the main purpose of MosquIoT and its associated architecture (Fig. 2) is to digitize the currently used traditional entomological *Ae. aegypti* mosquito surveillance systems by implementing a comprehensive digital ecosystem that allows these insects' oviposition to be monitored in cities and peri-urban areas. Accordingly, we seek to ensure that vector monitoring and surveillance performed by IoT sensors (MosquIoT) set up at several locations of the city submit their readings several times a day, and that those in charge of the entomological surveillance may easily have a view of the current situation or risk map. In this respect, the architecture was conceived with the aim of making its application in the industry as affordable as possible, bridging the gap between the scientific and the industrial world, and considering all the digital services a subject-matter expert might need to make prompt and accurate decisions. As regards the architecture, an IoT platform was implemented to receive, store and display all the data from the several surveillance sites where MosquIoT will operate. The IoT platform used for this research study was ThingsBoard [46], which was selected for its architecture based on micro-services and an embedded MQTT (Message Queue Telemetry Transport) broker, which supports two-way communications from MosquIoT, when connected to a WiFi network. In this respect, two-way MQTT communication capabilities between MosquIoT and the IoT platform were also ensured, supporting standard telemetry and Remote Procedure Calls (RPC) to send commands or orders from the Web Dashboard (frontend of the IoT platform) to the MosquIoT devices set up at the surveillance sites. At certain surveillance settings in which WiFi connectivity is unavailable (such as peri-urban or rural areas), The Things Networks (TTN) middleware [28] was also integrated to support LoRaWAN communications. This network is part of the LPWAN (Low-Power Wide-Area Network) group [29], which has become increasingly popular in the IoT field because of its inherent low consumption and broad coverage range [30], [31]. For this research study, we implemented a LoRaWAN in the Australian frequency band (AU915: 915-928 MHz), in compliance with the radio spectrum standards issued by the Argentine federal telecommunications authority (ENACOM). The implementation of LoRaWAN requires several components, namely, the gateway in charge of spreading the network, establishing a two-way communication with this network, and acting as a bridge to the TTN middleware. The gateway used in this study was Dragino LG02, which was integrated into TTN by means of the HTTPS (Hyper Text Transfer Protocol Secure) protocol. The TTN middleware was integrated into the IoT platform through MQTT over TLS v1.2, following the same logic as in WiFi communications. We also implemented two-way communications with field devices. Finally, the architecture also showcases the existence of a tool designed and developed for the system installer on the field—a Mobile App called "MosquIoTNet" intended to facilitate the installation, calibration and setup of MosquIoT at each surveillance site.

## IV. MOSQUIOT IMPLEMENTATION

Following the same methodology used in traditional ovitraps, each MosquIoT comprises a black recipient (Fig. 2) and has the same functional features as ovitraps, but with certain digital functionalities that will later be discussed in further detail. The MosquIoT recipient contains tap water with a certain amount of brewer's yeast to attract female mosquitos to these traps to lay their eggs, a disposable tongue depressor where mosquito eggs will be laid, and the MosquIoT's own electronics within a highly leak-proof case. In this respect, the primary purpose of MosquIoT is to automate the detection and count of *Ae. aegypti* eggs laid on the tongue depressor and to remotely send that information to the surveillance center. In determining the number and types of eggs laid, the MosquIoT sensor uses a camera as the primary sensor to detect the presence of eggs on the tongue depressor located inside the recipient. With this set of technologies, the purpose of the research is to design and develop an automatic and stand-alone system to sense and transmit data on the detected density of mosquitos and to send such telemetric data to the IoT platform that will automatically build the risk maps. MosquIoT was designed under a low-cost perspective, as opposed to the technological solutions referred to in the introduction to this paper [15], [14], [16], [18], [19], [20], [21], [22], which propose more expensive, complex and barely scalable alternatives. The construction of the first MosquIoT prototype required an investment of just USD 150. In a mass production setting, this cost might decrease by approximately 20%.

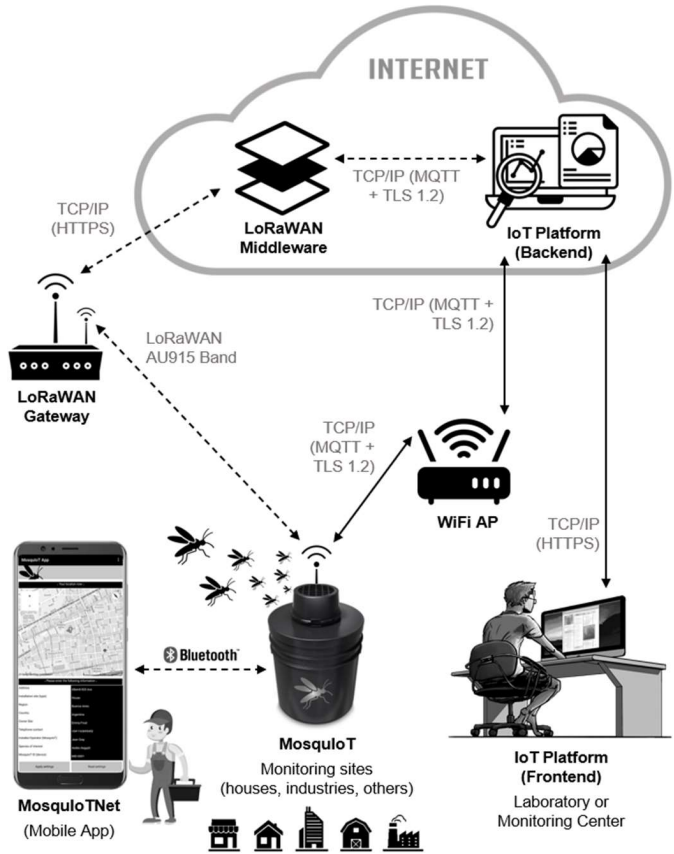

Fig. 2. Digital ecosystem associated with MosquIoT.

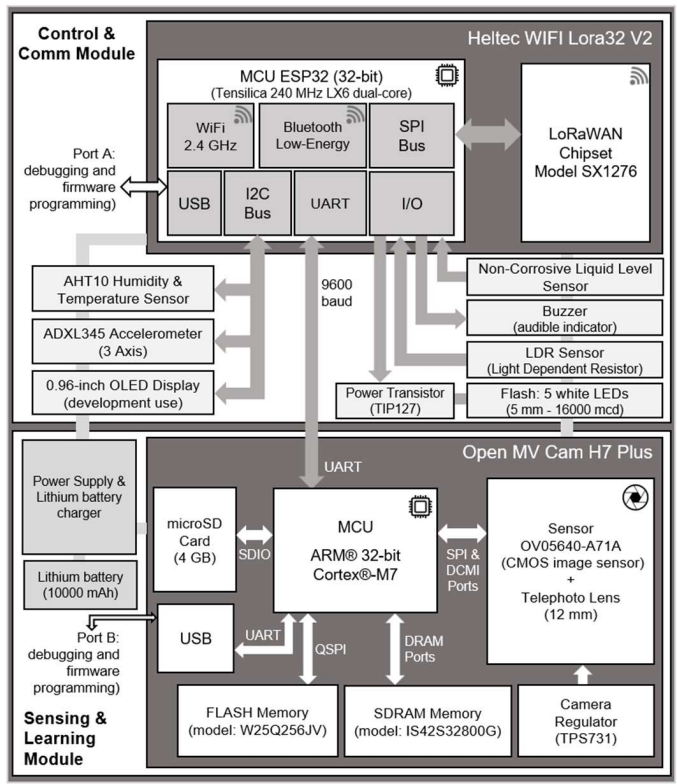

Fig. 3. Implemented hardware.

MosquIoT has two well-differentiated hardware modules (Fig. 3)—the Sensing & Learning module, in charge of detecting and counting *Ae. aegypti* eggs on the tongue depressor, and the Communications & Control module, which serves as the controller of the Sensing & Learning module and integrates the IoT communication engines (WiFi, LoRaWAN and BLE). In addition, the latter includes a set of sensors to ensure accurate system functioning in general: an environment temperature and humidity sensor, a light intensity sensor within the recipient, a 3-axis accelerometer that warns about the recipient potentially tipping over, and a water level sensor to detect the presence of water inside the recipient (without water, females will not lay their eggs).

*A. Sensing & Learning Module*

The Sensing & Learning module comprises a development board called OpenMV CAM H7 PLUS [32] and its add-ons: a telephoto lens and a 4 GB microSD (Micro Secure Digital) card to store captured snapshots, the firmware developed and supplemental libraries. The Sensing & Learning module also includes the hardware associated with the MosquIoT power supply—ability to connect an external 12 Volts DC (Direct Current) power supply and an internal power supply by means of a 10,000 mAh lithium battery and its respective charge and control electronic circuit. The battery is rechargeable (fast charge) and easily replaceable by the installer at the time of servicing (the time at which the water and tongue depressor are replaced). The OpenMV development board is a stand-alone artificial vision platform that features a camera and a powerful MCU (microcontroller). The MCU is an STM32H743II ARM

Cortex M7 running at 480 MHz with 32 MBs SDRAM, 1 MB of SRAM, and 32 MB of external flash + 2 MB of internal flash. The OpenMV board also features an OV5640-A071A CMOS image sensor [33] manufactured by OmniVision Technologies, capable of capturing 2592 x 1944 pixel (5 megapixel) images. For this study, the camera was set at a high definition (HD) of 1280 x 720 pixels. However, for the specific purposes of this study, a 512 x 512-pixel sub-resolution was established. As mentioned, a 12 mm telephoto lens (F2.0 opening and 1/3" format) was embedded to ensure the *Ae. aegypti* eggs are more accurately detected, read and counted.

*Learning Phase*

The recent developments in MCU architecture and the design of algorithms have enabled the execution of sophisticated Machine Learning workloads, even in small MCUs. Embedded Machine Learning— also known as TinyML [34], [35], [36], [37]—is the Machine Learning field used when applied to embedded systems such as that discussed here. TinyML can be implemented in low-power systems to perform complex automated tasks. There are significant benefits associated with the implementation of TinyML in IoT-based devices. Such benefits are clearly summarized by the BLERP acronym (Bandwidth, Latency, Economics, Reliability, and Privacy) [38], which means that the wideband of Machine Learning algorithms may retrieve meaningful information from data that would otherwise be inaccessible due to wideband limitations. As to latency, the Machine Learning models in the device are able to respond to inputs in real time, enabling applications as stand-alone vehicles that would otherwise not be feasible if they relied on the network latency. Concerning economics, since data are processed on the device (edge), embedded Machine Learning systems avoid the costs of transmitting data through a network and processing them in the cloud. In terms of reliability, systems controlled on the device by models of this kind are more reliable than those using a cloud connection. Finally, in terms of privacy, when data are processed within an embedded system and are never transmitted to the cloud, user privacy is secured.

In order for MosquIoT to be able to detect and quantify *Ae. aegypti* eggs, a TinyML model has to be designed, developed and implemented within the system. In this respect, for the purposes of this study, the selected open-source platform was Edge Impulse's Machine Learning [39], [40], [41], [42], [43], which supports TensorFlow Lite [44], [45] and enables the execution of customized imaging segmentation and classification models within IoT devices. In this study, we used the deep learning architecture customized by Edge Impulse known as FOMO (Faster Objects More Objects) [47], [48], [49] applicable to TinyML solutions. This architecture enables object detection in low computational capacity devices. The idea behind FOMO is finding the right balance between accuracy, speed, and memory to reduce learning models to extremely reduced sizes. The granularity of FOMO's output can be configured based on the application and can detect many instances of objects in a single image. In this respect, apart from FOMO, there are other traditional models such as YOLOv5 (You Only Look Once) [50] or MobileNet-SSD [51], but these are not recommended to detect many tiny objects of similar size within the same image. FOMO is based on the customization of MobileNetV2 [48], a convolutional neural network (CNN) developed by Google [53] released as part of the TensorFlow Lite image classification library. The MobileNetV2 architecture [52] promotes the development of technology applied to visual object recognition, including semantic classification, detection and segmentation. This article will not delve into the operation of MobileNetV2 but, in order to understand the architecture implemented in this work, is should be highlighted that it consists of the so-called "Depthwise Separable Convolutions" formed by a deep convolution, which filters the input and then applies a convolution with a kernel set to a size of 1x1 (convolution point). MobileNetV2 evolved to a block composed of three serial convolutions [52], [62], [63], [66], where the point convolution at the end of the block does just the opposite (reduces the number of channels), and the first layer of the block will now be an expansion layer with the name of "Bottleneck" that makes smaller the images that traverse the network. In FOMO (MobileNetV2) Rectified Linear Unit 6 (or ReLU6) [67] is used as the activation function and the output of the "Bottleneck" layer does not have the activation function applied. This layer produces low-dimensional data and the existence of a non-linearity after this layer would destroy useful information [62]. The parameters configured within FOMO (MobileNetV2) for the use case of interest are presented in Table I [48]. The mentioned features of MobileNetV2 make it performance-efficient [64], [65], which is the main reason why Edge Impulse developers it adopted and customized it for FOMO. The latter, in addition to having inherited functionalities from MobileNetV2, incorporates as well its own distinctive functionality that allows object detection based on input image analysis through grid division and heat maps [48]. A technique similar to image classification is then executed to classify the grids independently. By default, the grid size is 8×8 pixels. This means that for the sub-resolution established in the snapshots captured by MosquIoT (512×512 pixels), the heat map will be 64×64, making FOMO flexible and useful, even if images are large and objects are tiny. Instead of detecting bounding boxes as YOLO v5 and MobileNet-SSD do, FOMO predicts the center of the object, since many object detection solutions are only concerned with the location of the objects in the frame, rather than on their size. The centroid detection feature used by FOMO is more efficient from a computing perspective than the prediction of the bounding box and requires less data or parameter-settings [48]. In summary, FOMO is an algorithm that allows objects to be detected, monitored and counted in real time, and which may be run locally in MCUs, thanks to the possibility of using its SDK (Software Development Kit) in MicroPython [54].

For this work, the Machine Learning algorithm construction was based on 4 building blocks: input, processing, learning, and deployment. The input block involved defining the type of data to be used for model training. In this case, the input data were the snapshots of the tongue depressors and their associated eggs. The processing block is a feature extraction system based on Digital Signal Processing (DSP) operations from which the model learns [55] and standardizes the images, converting each pixel channel into a floating value between 0 and 1. The deep learning model (FOMO) was then parameterized and implemented. This model allows information on the object of interest (*Ae. aegypti* eggs) to be generated, determining quantities and positions in the plane of the objects of interest observed in the image. The learning block was set up with an 0.80 threshold, which is the minimum confidence level required to ascertain the presence of *Ae. aegypti* eggs on the tongue depressors. In addition to the aforementioned threshold, general training adjustments were also made within the Edge Impulse platform, which are presented in Table II [68]. Once the learning algorithm was built, the model was deployed within MosquIoT (this process will be described in Section IV-C).

TABLE I
FOMO (MOBILENETV2) SETTINGS

| Parameter / Field | Content |
|---|---|
| layers | 1 |
| filters | 32 |
| kernel_size | 1 |
| strides | 1 |
| activation | ReLU6 |

TABLE II
GENERAL TRAINING SETTINGS

| Parameter / Field | Content |
|---|---|
| training_model | keras |
| alpha | 0.35 |
| training_cycles | 200 |
| learning_rate | 0.001 |
| input_num_channels | 1 |
| disable_per_channel_quantization | false. |
| num_classes | 1 |
| train_dataset | 90 |
| override_mode | segmentation |
| profile | quatized (in8) |

*B. Communications & Control Module*

From the point of view of ensuring the automatic and autonomous operation of MosquIoT (at each monitoring place), the Communications & Control module plays a very important role, since it provides edge intelligence that will ensure that monitoring is carried out at each with efficiency. In this context and in the event of an operational problem, MosquIoT is expected to issue alerts in real time to those responsible for monitoring to carry out the corresponding corrective actions (as explained in Section I, please consider that this does not occur in current traditional systems). To meet this requirement, the sensors integrated were the following: an accelerometer [70], which informs in real time about the state of the physical position of the recipient (it was customized in such a way that it only communicates two possible states: well-positioned and overturned), a Light-Dependent Resistor (LDR) [71] to determine if the lid of the recipient was opened (if this happens, oviposition will not take place), a non-corrosive liquid level sensor to detect the presence or absence of water inside the recipient [73] (if there is no water, oviposition will not occur) and finally, a power control module [72] was incorporated, made up of an array of high-brightness LEDs that allow the tongue depressor to be illuminated as well as the associated eggs (this ensures the homogeneity of the measurements by the main MosquIoT sensor). Additionally, in the Communications

& Control module, a temperature and relative humidity sensor [74] was incorporated. It provides specific meteorological information from the place where they were detected and is complemented with the information sent about the number of eggs detected. This sensor provides valuable information for expert entomological monitoring personnel about oviposition trends and probabilities at monitoring places. The architecture of the MosquIoT Communications & Control module was designed in such a way that it allows it to fulfill the previously mentioned functions and, from a more technical aspect, its objectives were oriented towards centralized computing, control of IoT communication links and the possibility of having robust native electronic interfaces (I/O) to integrate plug-in sensors and actuators. To meet these requirements, it was decided to integrate the Heltec WiFi Lora32 V2 board [57], which has a powerful 32-bit dual-core MCU based on ESP32 Tensilica 240 MHz LX6 [58]. In addition to this important computing resource, the Heltec board integrates the SX1276 (915 MHz) [69] chipset for LoRaWAN communications, a 2.4 GHz WiFi module and a BLE module to interact bidirectionally with MosquIoTNet (Mobile App).

### C. Firmware Developed

As mentioned in Section IV-B, the Communications & Control module includes a powerful MCU, thanks to which we were able to design, develop and express the system's operating logic, using complex firmware with the C/C++ programming language. Arduino and Visual Studio Code Integrated Development Environments (IDEs) were used to develop the firmware. "ESP-Prog" hardware was used to debug this firmware, which is connected to the ESP32 JTAG (Joint Test Action Group) port for real-time debugging.

In terms of IoT communications, the firmware of this module is able to connect and operate in WiFi networks (fixed or mobile). For this communication link, the MQTT protocol (over TLS 1.2) and a RPC-based protocol were implemented in the firmware, allowing information and orders from the IoT platform's Web Dashboard (frontend) to be received. MosquIoT is also able to connect to LoRaWAN networks. Furthermore, the mechanisms required to send (Uplink) and receive (Downlink) messages through these LPWAN networks were implemented in the firmware. In the latter case, to receive information or orders from the IoT platform's Web Dashboard (e.g., remote eggs reading order on demand or rescheduling at the event transmission frequency). Due to the reduced payload size defined by the LoRaWAN standard for data messaging, specifically for the frequency schedule established and governed by ENACOM (AU915), the defined Payload ranges from 11 to 242 bytes. The Cayenne Low Power Payload (LPP) [59] library was used, which provides a convenient and easy way to send data over LPWAN networks such as LoRaWAN. Cayenne LPP is a standardized and proven format that enables a substantial reduction in the number of bytes to be sent and allows MosquIoT to simultaneously send multiple data from its sensors and context parameters, based on pre-established rules and technical characteristics. In our specific case, the use of Cayenne LPP helped reduce the payload to 110 bytes, including the Cayenne global structure (data ID + data type + data size).

In order to set up MosquIoT (at the surveillance site), the installer uses the MosquIoTNet App to send the applicable configuration for it to be able to operate autonomously and automatically. In this respect, MosquIoT firmware administers and saves such configurations on its Flash memory (non-volatile). MosquIoT firmware also manages its Static Random-Access Memory (SRAM) for certain technical and functional circumstances in which the information is used and protected in a volatile manner. Upon completion of MosquIoT configuration, the installer may request a reading from the MosquIoTNet App. The Control & Communications module firmware requests a reading from the Sensing & Learning module through the Universal Asynchronous Receiver-Transmitter (UART) port. Ten seconds later—the average time it takes for the sensor (camera) to send the response—the firmware processes such information and sends the results to the App to notify the user. A few seconds later, MosquIoT submits all the information related to the test to the IoT platform through previously established connectivity. Fig. 4 shows a Unified Modeling Language (UML) sequence diagram, a conceptual scheme that expresses the behavior and interaction of the MosquIoT firmware with the other components of the system. MosquIoT firmware also features programmable functional logic for event transmission time frequency to the IoT platform and prior to such transmission. MosquIoT firmware reads the tongue depressor to check for the presence of *Ae. aegypti* eggs. The time frequency of the reading set in MosquIoT depends on the communication link established and the battery autonomy. In case of establishing a WiFi link, and in order for the battery to last 7 days, transmission frequency is recommended to be set up once daily (under any circumstances, the sole event submitted will have 4 readings performed every 6 hours, with its pertinent timestamp). In case of programming the LoRaWAN communication link for 7 days' autonomy, the recommended transmission frequency is 4 times a day.

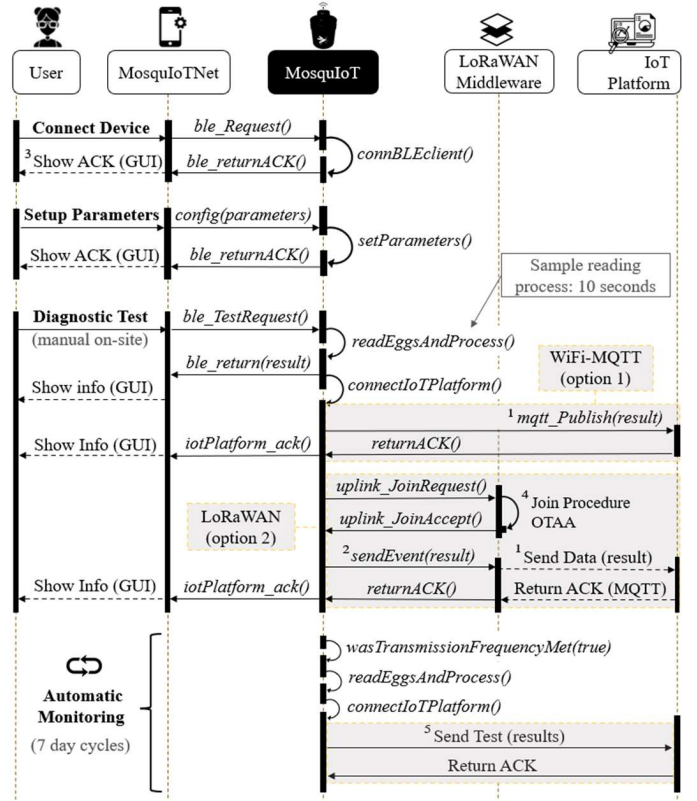

Fig. 4. UML sequence diagram of the firmware (MosquIoT)

The Sensing & Learning module encompasses the design and development of a complex firmware program using MicroPython programming language, and OpenMV LLC's OpenMV IDE (v2.9.0) as programming IDE [60]. A USB port with an OpenMV board was used to debug the firmware developed. In addition, within the same IDE, Serial Terminal, RGB frame buffer viewer (Red, Green and Blue) and Histogram Display were used for debugging, displaying in real time all processes run by the system's operating system, the performance of the firmware developed, and real-time images of the camera showing the objects of interest being read and quantified. The firmware developed detects and quantifies the *Ae. aegypti* eggs: in order to initially perform this task, the firmware takes 5 snapshots every 2 seconds, and prompts a function based on the trained and optimized model for *Ae. aegypti* egg reading (described in Section IV-A). When the model detects the presence of eggs of the species of interest, it assigns an individual ID to each, globally quantifies them, and finally prompts a function that calculates the individual confidence score of each egg read. The latter function also averages the individual confidence score of each of the 5 snapshots. If that Score is equal to or higher than 0.80, the eggs will be deemed to belong to the *Ae. aegypti* species. To complete the process, when the reading results are available, such information is communicated through the UART port, reporting the number of eggs read, timestamp and context data inherent to the OpenMV camera (sensor model, hardware version, firmware version, HAL or Hardware Abstraction Layer version, operating system, frame size and pixformat). In addition, the firmware also saves all captured snapshots in a microSD memory so as to have evidence and traceability of the readings performed.

## V. MOSQUIOTNET

During the design and development process of MosquIoT, the creation of a Mobile App (MosquIoTNet) was considered essential for the installation, configuration, calibration and functional diagnosis of the device at the surveillance site. MIT App Inventor [61] was the tool selected to develop the App.

When the user runs MosquIoTNet for the first time, it will be prompted to enable the use of two specific resources: BLE and GPS. These resources are essential to communicate MosquIoTNet with MosquIoT (through BLE) and for the App to be able to transfer the GPS coordinates to the device, for it to be subsequently geo-referenced from the IoT platform. From a functional perspective, MosquIoTNet allows the user to set up the required parameters to ensure the successful and autonomous operation of MosquIoT at the surveillance site. Fig. 5 presents a UML sequence diagram, representing these functionalities. The setup parameters to be input into MosquIoTNet include the address, province and country where the device is running, person in charge of the site (name and contact details), type of place where MosquIoT is running (home, business, public building, factory or field), name of installer in charge (member of the mosquito surveillance campaign staff), species of interest (in this case, *Ae. aegypti*), device ID previously assigned from the IoT platform, according to the surveillance campaign established by the experts and, finally, the possibility to enable and set up MosquIoT's IoT communication engine. Concerning the last of these, the WiFi network parameters can be set up from MosquIoTNet (establishing the network and password). If the device is located in a place with LoRaWAN coverage, such connectivity may be easily established from the App.

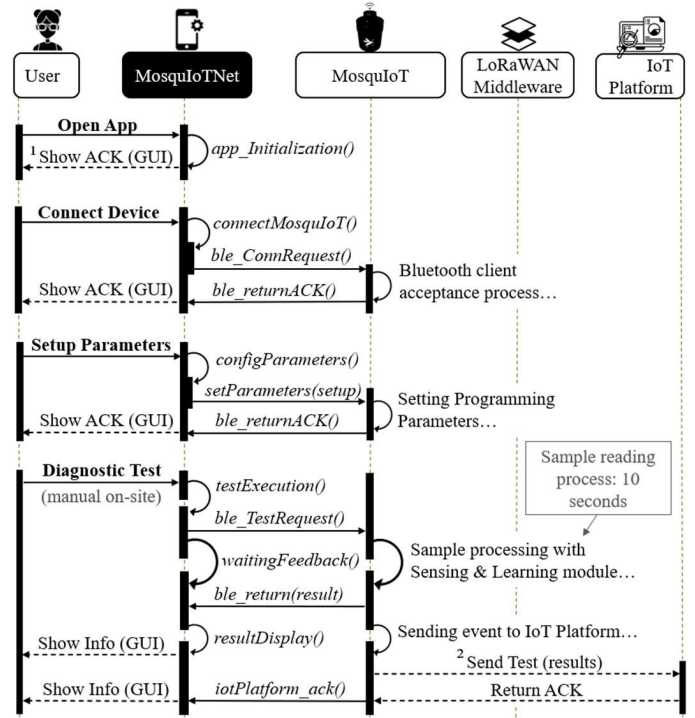

¹ GUI (Graphical User Interface), ² MosquIoT Connectivity (options): 1 → WiFi-MQTT or 2 → LoRaWAN

Fig. 5. UML sequence diagram of the MosquIoTNet (Mobile App)

When setting up MosquIoT (at the surveillance site), local diagnosis tests may be carried out from MosquIoTNet to calibrate the device and request readings on demand to rapidly calculate the number of eggs read. When the user requests a test reading from MosquIoT, the system will submit the results within approximately 10 seconds, displaying on MosquIoTNet the number of eggs read, timestamp and an ID the assay. Finally, MosquIoTNet also notifies the user if the data have been successfully transmitted to the IoT platform.

## VI. IOT PLATFORM

ThingsBoard [46] is the IoT platform selected for this study, as it allows generic and customized IoT devices to be collected, processed, visualized and managed. The platform combines scalability and performance to ensure data persistence. ThingsBoard enables the secure provision, supervision and monitoring of IoT entities, through enriched Application Programming Interfaces (APIs) from the server side. The platform is able to transform and standardize data from the devices, and set up and trigger incoming telemetry event alarms, attribute updates, device downtime, and users' actions.

A Web Dashboard (frontend) was designed and developed within ThingsBoard, which allows the surveillance of each MosquIoT device installed to be visualized in real time (whether transmitting from a WiFi network or LoRaWAN). In this respect, the user may have access to information specific to each installed device: MosquIoT geo-location through interactive maps, the address, province and country in which the device is running, person in charge of the site (name and contact details), type of place where MosquIoT is running, and name of installer in charge. In addition to the context information, the Web Dashboard displays information about the surveillance status at each specific site, detailing the species and number of mosquito eggs read in the last communication to the platform. The Web Dashboard also displays specific technical information about MosquIoT, such as information on the status of communications (type of communication link, signal level, and types of events submitted), and specific MosquIoT data (device ID and firmware version). It also provides supplemental data about the main egg reading sensor (in this case, the OpenMV CAM): sensor name, hardware version, firmware version, HAL version, and specific camera configuration details (operating system version, frame size and pixformat). In addition, the platform also displays the temperature and humidity details typically transmitted by MosquIoT. Default rules were defined in the platform for these physical quantities to establish maximum and minimum thresholds. Based on these thresholds, e-mail notifications or alerts are sent when such quantities infringe the rules. Looking forward, this functionality seeks to maximize egg reading success on site and provide the platform with basic data to train Machine Learning modules for users to be able to better understand these vectors at the specific surveillance sites. In addition, as explained in previous sections, MosquIoT features an accelerometer to pinpoint the device position and notify the platform upon detecting the permanent tilt of the device. The platform displays information on MosquIoT's position and is able to create a rule to trigger and send an e-mail alert to the person in charge of the site upon receiving a tilt event. When receiving an event of this nature, it is understood that MosquIoT was tilted, that the water inside it was spilled and, consequently, that the samples contained in the device will no longer be useful for entomological surveillance. In this respect, the person in charge of the site will have to take corrective actions in situ. Finally, the Web Dashboard also displays the water level status inside MosquIoT (a vital input for *Ae. aegypti* to lay eggs). All information submitted from MosquIoT persists on databases within the IoT platform and is available for the user to perform searches by date and time, ensuring the respective system traceability. Additionally, the IoT platform has a REST (Representational State Transfer) API, which allows all retrieved data and their time series to be accessed and a copy of such data be rapidly obtained for subsequent processing with another third party's platforms to perform future studies.

## VII. EXPERIMENTAL VALIDATION

Most Machine Learning projects begin by collecting data. For the purposes of this study, we obtained a substantial amount of *Ae. aegypti* eggs on the field by means of traditional ovitraps. With these samples, a training and validation plan was designed to develop the Machine Learning algorithm that would allow MosquIoT to recognize and quantify *Ae. aegypti* eggs: a total of 90 tongue depressors with 715 *Ae. aegypti* eggs were established for the training samples, while 10 tongue depressors with 67 eggs were established for the validation set. The eggs on all the tongue depressors (100) were quantified using the same methodology implemented by labs or subject-matter experts: a manual reading of each tongue depressor and its associated eggs was performed by means of a TMPZ-C1200 Galileo microscope. Once the eggs on each tongue depressor (training and validation) were quantified, a firmware program was developed in MosquIoT to capture snapshots at each of the 90 tongue depressors set for the training model (explained in Section IV-C). These snapshots were taken at an HD sub-resolution of 512 x 512 pixels (in which MosquIoT will operate in the practice). By way of example, Fig. 6 shows one of the 90 tongue depressors used for model training purposes (tongue depressor 16 with 6 *Ae. aegypti* eggs).

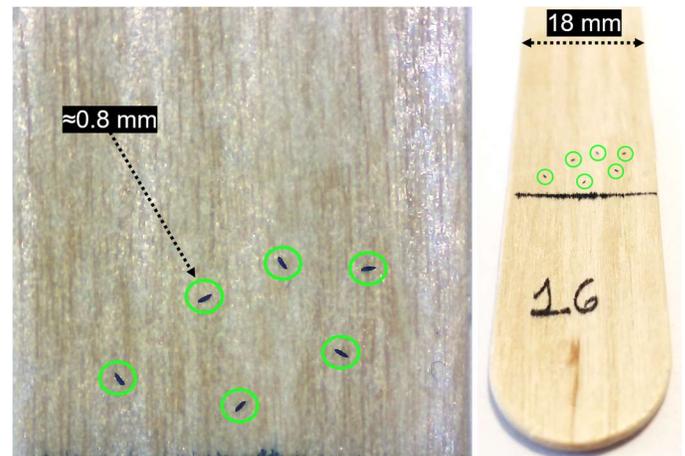

Fig. 6. Tongue depressor N°16 used for model training.

Once the Machine Learning model was implemented within MosquIoT, actual assays were performed with the validation set (10 tongue depressors with 67 eggs). Validation assays were performed in a lab, using a customized MosquIoT prototype to be able to operate in this type of setting (Fig. 7). As part of the validation process performed, apart from *Ae. aegypti* eggs, other objects of varying sizes, colors and textures (seeds, soil, stones, and grains) were laid on tongue depressors in an attempt to "mislead" the model developed and implemented in MosquIoT in order to validate the model robustness. By way of example, Fig. 8 shows tongue depressors (samples 02, 05, 06 and 07) with their respective validation results (the green circle is indicative of the successful detection of eggs). It is worth mentioning that in the validation assays performed, no eggs from species other than *Ae. aegypti* were used. This is because, in the geographic area of the study (central region of Argentina), no other species of mosquitoes lay their eggs on artificial recipient walls. Therefore, there are no possibilities for MosquIoT to be "misled" by eggs from other species on the tongue depressors. The only mosquito species ovipositing on the ovitraps are *Ae. aegypti* and *Ae. albopictus* [7]. The latter is not present in the region of interest [56].

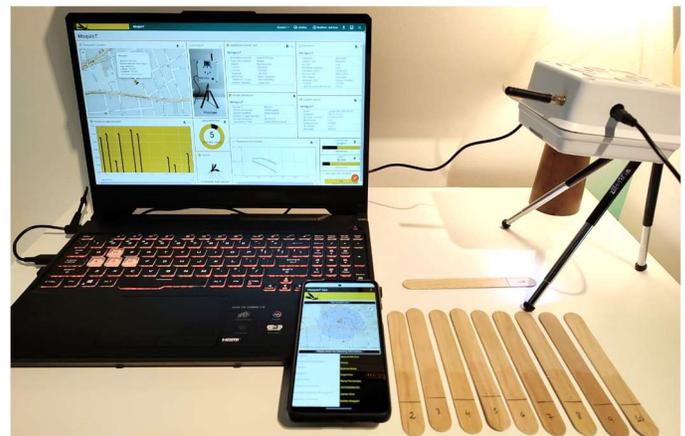

Fig. 7. Experimental setup

The results obtained were highly promising, in that MosquIoT succeeded in detecting and quantifying all the eggs on the 10 tongue depressors designated as the validation set. Table III shows the results obtained, which are arranged as follows (from left to right): ID of each tongue depressor used for validation assays (ID Sample), numbers of eggs read and counted manually using the lab microscope, number of eggs read and counted automatically by MosquIoT, and ID of each egg read and counted by MosquIoT (Egg ID). The last column shows the average individual confidence score for each egg read by MosquIoT. In order for an object read by MosquIoT to be considered an *Ae. aegypti* egg, it should be equal to, or higher than, 0.80, as can be seen from the analysis performed and averages calculated by the MosquIoT algorithm when taking 5 snapshots with the same sample. Each of the 67 eggs read by MosquIoT exceeded the established threshold. The model validation process was thus successful.

Once the tests were carried out with the validation samples, focused on the accuracy of the TinyML model implemented within MosquIoT, a global test (end-to-end) was carried out to validate the correct functionality of all the digital components in an automatic and autonomous format: from the sensor (MosquIoT) to the IoT platform through the WiFi communication link (transmitting 4 times a day, frequency of 6 hours). Let us highlight that the time segmentation responds to the current and traditional sample collection procedures carried out by the personnel assigned to the entomological campaigns to monitor *Ae. aegypti* (explained in Section I of this work).

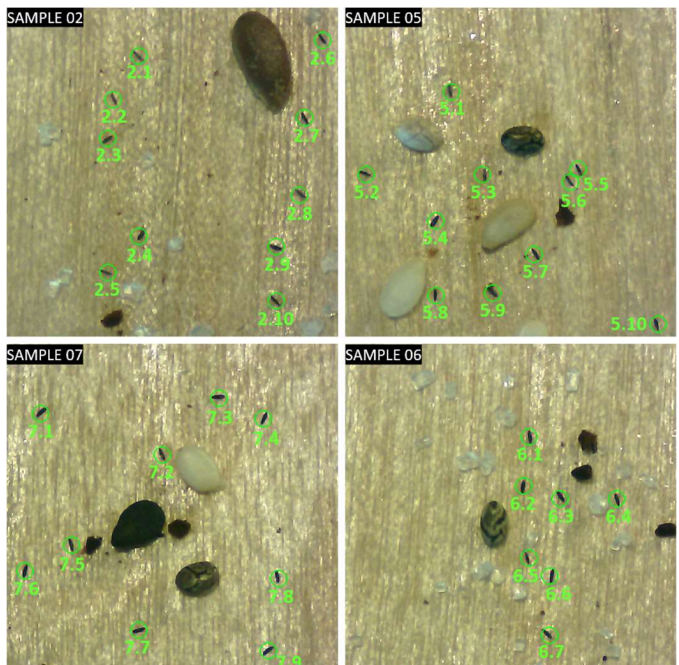

Fig. 8. Tongue depressors used to validate the trained model.

TABLE III
RESULTS FROM THE TEST SAMPLES

| Tongue depressor (ID Sample) | Number of existing Eggs [1] | Numbers of Eggs read with MosquIoT | Egg ID | MosquIoT Results (confidence score per Egg read) [2] |
|---|---|---|---|---|
| 01 | 3 | 3 | 1.1 | 0.98 ✓ |
| | | | 1.2 | 0.95 ✓ |
| | | | 1.3 | 0.97 ✓ |
| 02 (Fig. 8) | 10 | 10 | 2.1 | 0.98 ✓ |
| | | | 2.2 | 0.94 ✓ |
| | | | 2.3 | 0.99 ✓ |
| | | | 2.4 | 0.98 ✓ |
| | | | 2.5 | 0.84 ✓ |
| | | | 2.6 | 0.87 ✓ |
| | | | 2.7 | 0.94 ✓ |
| | | | 2.8 | 0.98 ✓ |
| | | | 2.9 | 0.95 ✓ |
| | | | 2.10 | 0.94 ✓ |
| 03 | 2 | 2 | 3.1 | 0.95 ✓ |
| | | | 3.2 | 0.93 ✓ |
| 04 | 8 | 8 | 4.1 | 0.87 ✓ |
| | | | 4.2 | 0.88 ✓ |
| | | | 4.3 | 0.99 ✓ |
| | | | 4.4 | 0.98 ✓ |
| | | | 4.5 | 0.97 ✓ |
| | | | 4.6 | 0.99 ✓ |
| | | | 4.7 | 0.98 ✓ |
| | | | 4.8 | 0.90 ✓ |
| 05 (Fig. 8) | 10 | 10 | 5.1 | 0.98 ✓ |
| | | | 5.2 | 0.91 ✓ |
| | | | 5.3 | 0.99 ✓ |
| | | | 5.4 | 0.89 ✓ |
| | | | 5.5 | 0.96 ✓ |
| | | | 5.6 | 0.99 ✓ |
| | | | 5.7 | 0.96 ✓ |
| | | | 5.8 | 0.82 ✓ |
| | | | 5.9 | 0.97 ✓ |
| | | | 5.10 | 0.80 ✓ |
| 06 (Fig. 8) | 7 | 7 | 6.1 | 0.99 ✓ |
| | | | 6.2 | 0.94 ✓ |
| | | | 6.3 | 0.95 ✓ |
| | | | 6.4 | 0.99 ✓ |
| | | | 6.5 | 0.82 ✓ |
| | | | 6.6 | 0.98 ✓ |
| | | | 6.7 | 0.90 ✓ |
| 07 (Fig. 8) | 9 | 9 | 7.1 | 0.97 ✓ |
| | | | 7.2 | 0.90 ✓ |
| | | | 7.3 | 0.93 ✓ |
| | | | 7.4 | 0.91 ✓ |
| | | | 7.5 | 0.83 ✓ |
| | | | 7.6 | 0.98 ✓ |
| | | | 7.7 | 0.94 ✓ |
| | | | 7.8 | 0.86 ✓ |
| | | | 7.9 | 0.98 ✓ |
| 08 | 4 | 4 | 8.1 | 0.90 ✓ |
| | | | 8.2 | 0.99 ✓ |
| | | | 8.3 | 0.88 ✓ |
| | | | 8.4 | 0.94 ✓ |
| 09 | 5 | 5 | 9.1 | 0.93 ✓ |
| | | | 9.2 | 0.89 ✓ |
| | | | 9.3 | 0.97 ✓ |
| | | | 9.4 | 0.82 ✓ |
| | | | 9.5 | 0.99 ✓ |
| 10 | 9 | 9 | 10.1 | 0.91 ✓ |
| | | | 10.2 | 0.93 ✓ |
| | | | 10.3 | 0.84 ✓ |
| | | | 10.4 | 0.92 ✓ |
| | | | 10.5 | 0.97 ✓ |
| | | | 10.6 | 0.98 ✓ |
| | | | 10.7 | 0.85 ✓ |
| | | | 10.8 | 0.91 ✓ |
| | | | 10.9 | 0.95 ✓ |

[1] Readings were taken through a laboratory microscope.
[2] Each result expressed is the result of an average of 5 snapshots made with MosquIoT

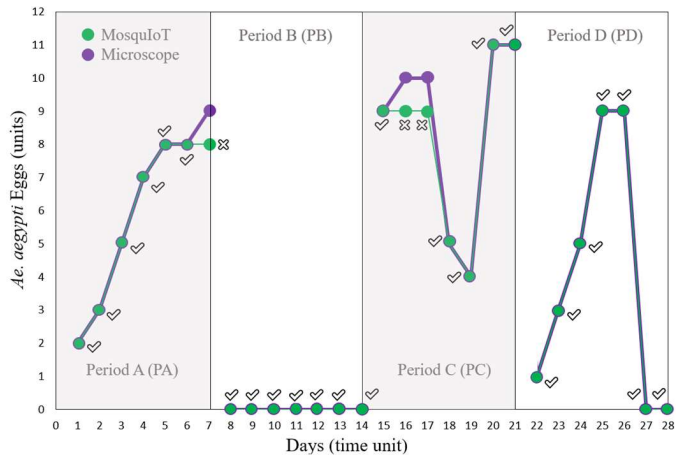

Fig. 9. Results of the Proof of Concept carried out during 28 days.

For this PoC, one tongue depressor was prepared and installed per day with a variable number of *Ae. aegypti* eggs that responded to the possible behaviors that could be manifested in the field. In the first period (PA) of the PoC, a tongue depressor was installed that increased the number of eggs sequentially as the days passed (typical and expected behavior of the mosquito). In the second period (PB), there was no presence of eggs (a behavior that could also occur). In the third period (PC), there was a behavior of egg loss and with the passing of days, a significant increase in oviposition. In the last period (PD), a behavior of egg loss due to flooding or overturning of the recipient was noticed (Fig. 9).

The results obtained from the PoC were very promising since the classification accuracy of MosquIoT was ≈ 97.67 %, managing to identify and count 126 of the 129 eggs that were and counted manually with the Galileo TMPZ-C1200 laboratory microscope (Table IV). From the point of view of the performance and stability of the MosquIoT firmware, during the 672 hours of continuous operation that the PoC required, it did not show any stability problems, which successfully complied with its routine of sensing tongue depressors every 6 hours during the 28 test days. From the point of view of stability in terms of communications, MosquIoT carried out 112 communications in a timely manner through a WiFi network to the IoT platform. In addition, it was possible to verify that there were no lags greater than 1 second between the MosquIoT timestamp and the one of timestamp of the event received in the MQTT broker of the IoT platform. To conclude, the first MosquIoT laboratory prototype achieved important and promising results that will enable its evolution in the next stages of the project.

## VIII. Conclusions

This work concerns the design, development and functional testing of MosquIoT, a device based on traditional ovitraps powered by IoT and TinyML technologies, digitizing and automating the reading and counting of eggs laid by *Ae. aegypti* —the vector mosquito of the dengue and other viruses. This system could prove extremely useful to the provincial and municipal authorities of Argentina (geographic area of application) to have a dynamic understanding of the populations of this insect and to prevent diseases, such as dengue, chikungunya, Zika and others.

The major benefit of MosquIoT within the current traditional ovitrap context lies in the proposed technology performing automatic and ongoing surveillance (once or several times a day). Moreover, thanks to the IoT platform implemented, the system is able to build risk maps to provide the authorities with a dynamic surveillance tool in real time. In traditional ovitraps, the process is slow and manual. Egg reading or detection is performed manually with the aid of a laboratory microscope 7 days after the installation of the recipient at the surveillance site (this methodology is more corrective than preventive and human error could be frequent). This work introduces the first MosquIoT laboratory prototype which, through evidenced training and validation processes, has been proven to work efficiently and effectively, from reading and counting eggs, to submitting and displaying the information on the IoT platform's Web Dashboard (frontend). In addition, in order to successfully run MosquIoT installation, configuration and functional diagnosis processes at the various surveillance sites, the MosquIoTNet App was developed.

## TABLE IV
### PROOF OF CONCEPT RESULTS

| Tongue depressor (ID Sample) | Day | Period (7 days) | Numbers of Eggs read | | | | | Result |
| --- | --- | --- | --- | --- | --- | --- | --- | --- |
| | | | Laboratory Microscope | MosquIoT | | | | |
| | | | | M1 [1] | M2 [1] | M3 [1] | M4 [1] | |
| 1 | 1 | PA | 2 | 2 | 2 | 2 | 2 | ✅ |
| 2 | 2 | | 3 | 3 | 3 | 3 | 3 | ✅ |
| 3 | 3 | | 5 | 5 | 5 | 5 | 5 | ✅ |
| 4 | 4 | | 7 | 7 | 7 | 7 | 7 | ✅ |
| 5 | 5 | | 8 | 8 | 8 | 8 | 8 | ✅ |
| 6 | 6 | | 8 | 8 | 8 | 8 | 8 | ✅ |
| 7 | 7 | | 9 | 8 | 8 | 8 | 8 | ❎ |
| 8 | 8 | PB | 0 | 0 | 0 | 0 | 0 | ✅ |
| 9 | 9 | | 0 | 0 | 0 | 0 | 0 | ✅ |
| 10 | 10 | | 0 | 0 | 0 | 0 | 0 | ✅ |
| 11 | 11 | | 0 | 0 | 0 | 0 | 0 | ✅ |
| 12 | 12 | | 0 | 0 | 0 | 0 | 0 | ✅ |
| 13 | 13 | | 0 | 0 | 0 | 0 | 0 | ✅ |
| 14 | 14 | | 0 | 0 | 0 | 0 | 0 | ✅ |
| 15 | 15 | PC | 9 | 9 | 9 | 9 | 9 | ✅ |
| 16 | 16 | | 10 | 9 | 9 | 9 | 9 | ❎ |
| 17 | 17 | | 10 | 9 | 9 | 9 | 9 | ❎ |
| 18 | 18 | | 5 | 5 | 5 | 5 | 5 | ✅ |
| 19 | 19 | | 4 | 4 | 4 | 4 | 4 | ✅ |
| 20 | 20 | | 11 | 11 | 11 | 11 | 11 | ✅ |
| 21 | 21 | | 11 | 11 | 11 | 11 | 11 | ✅ |
| 22 | 22 | PD | 1 | 1 | 1 | 1 | 1 | ✅ |
| 23 | 23 | | 3 | 3 | 3 | 3 | 3 | ✅ |
| 24 | 24 | | 5 | 5 | 5 | 5 | 5 | ✅ |
| 25 | 25 | | 9 | 9 | 9 | 9 | 9 | ✅ |
| 26 | 26 | | 9 | 9 | 9 | 9 | 9 | ✅ |
| 27 | 27 | | 0 | 0 | 0 | 0 | 0 | ✅ |
| 28 | 28 | | 0 | 0 | 0 | 0 | 0 | ✅ |
| Totals & Accuracy | | | 129 | 126 | 126 | 126 | 126 | ≈ 97,67 % |

[1] Measurements made automatically by MosquIoT (frequency: 6 hs)

Concerning future works, and based on the promising results obtained from the first laboratory prototype, a next phase is envisaged involving a new Proof of Concept (PoC) with an upgraded version of MosquIoT fitted to outdoor settings, as well as field assays, deploying specimens in certain cities of the region of interest to visualize the respective risk map and the associated predictive model on the IoT platform. The primary purpose of the PoC is to show that, by making the current ovitrap system digital with MosquIoT, the current *Ae. aegypti* detection campaigns, which are intrinsically associated with the prevention of *Ae. aegypti* -borne diseases, can be enhanced and taken to the next level.

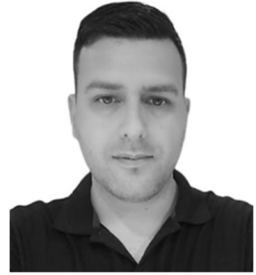

**Javier Aira** gained his bachelor's degree in management of automation and robotics systems at the National University of Lomas de Zamora, Buenos Aires, in 2016, and his master's degree in strategic management in information technology from the European University of the Atlantic, Santander, Spain, in 2020. He is currently pursuing a Ph.D. in advanced computer technologies in the IoT discipline in smart cities with the University of Castilla-La Mancha, Albacete, Spain. He has more than 18 years of experience in the industry, specifically in the area of R&D applied to fully digital solutions.

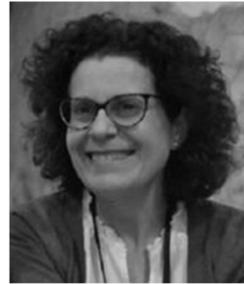

**Teresa Olivares** is an Associate Professor with the Department of Computing Systems at the University of Castilla-La Mancha. She received her Ph.D. degree in Computer Science in 2003 from the same university. She is a member of the research group High-Performance Networks and Architectures at the Albacete Research Institute of Informatics. Her main scientific research interests include Internet of Things standards, communications and protocols, heterogeneous low power wireless sensor networks and standards, smart environments, Industry 4.0 and Reverse Logistics. She has participated in more than 40 research projects and has co-authored more than 50 research papers in journals, conferences and book chapters.

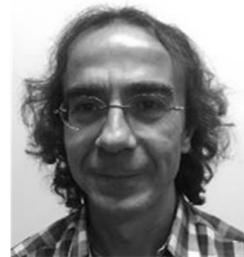

**Francisco M. Delicado** received his Ph.D. in Computer Engineering in 2005 from the University of Castilla-La Mancha. He has been an Associate Professor at the Department of Computer Engineering at this university since 2007 and member of the research group High-Performance Networks and Architectures at the Albacete Research Institute of Informatics. His research interests include SDN (Software Defined Networking), WSN (Wireless Sensor Networks), heterogeneous low power WSN and cloud networking.

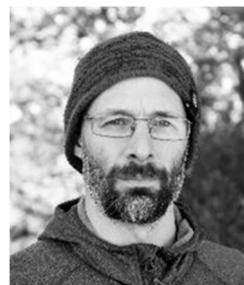

**Darío Vezzani** is member of the National Scientific and Technical research Council (CONICET) of Argentina. He received his Ph.D. in Biological Science in 2003 from the University of Buenos Aires. His research interests include the ecology and control of mosquitoes. He is currently working at the Ecosystems Institute at the National University of the Center of Buenos Aires Province, in Tandil City, Argentina.